\documentclass[letterpaper, 10 pt, conference]{ieeeconf}  

\IEEEoverridecommandlockouts                              

\usepackage{graphics} 
\usepackage{epsfig} 
\usepackage{amsmath} 
\usepackage{amssymb}  
\usepackage{subfig}
\usepackage{multirow}
\usepackage{url}

\title{\LARGE \bf
Generating Goal-Directed Visuomotor Plans Based on Learning Using a Predictive Coding-type Deep Visuomotor Recurrent Neural Network Model
}

\author{
Minkyu Choi$^{1}$, 
Takazumi Matsumoto$^{1}$, 
Minju Jung$^{1,2}$
and Jun Tani$^{1,*}$\\
\thanks{*Corresponding author}
\thanks{$^1$ Okinawa Institute of Science and Technology (OIST), Okinawa, Japan}
\thanks{$^2$ Korea Advanced Institute of Science and Technology (KAIST), Daejeon, Korea}
\thanks{\{minkyu.choi8904, takazumi, minju5436, tani1216jp\}@gmail.com}
}

\begin{document}

\maketitle
\thispagestyle{empty}
\pagestyle{empty}

\begin{abstract}

The current paper presents how a predictive coding type deep recurrent neural networks can generate vision-based goal-directed plans based on prior learning experience by examining experiment results using a real arm robot. The proposed deep recurrent neural network learns to predict visuo-proprioceptive sequences by extracting an adequate predictive model from various visuomotor experiences related to object-directed behaviors. The predictive model was developed in terms of mapping from intention state space to expected visuo-proprioceptive sequences space through iterative learning. Our arm robot experiments adopted with three different tasks with different levels of difficulty showed that the error minimization principle in the predictive coding framework applied to inference of the optimal intention states for given goal states can generate goal-directed plans even for unlearned goal states with generalization. It was, however, shown that sufficient generalization requires relatively large number of learning trajectories. The paper discusses possible countermeasure to overcome this problem.

\end{abstract}

\section{INTRODUCTION}

Recently robotics researchers have focused more on studies on learnable robot by using advanced schemes of deep learning \cite{Cangelosi,levineabbel,hwangsystem,ogata}. Obvious benefit is that learning by robots themselves can ease difficulty in describing precise models of the robots and their environment by the users. The most popular approach in applying deep learning to robots is to use convolutional neural network (CNN) for developing visuomotor mapping possibly by using reinforcement learning framework \cite{peter,levineabbel}. Another interesting approach is using the framework of predictive coding \cite{rao,friston} to robot learning problems \cite{book, taninolfi,hwangsystem,ogata,nagai,yamashita}. The predictive coding framework can allow robots to develop task specific internal models by extracting latent causality between intention states and the resultant outcomes of perceptual sequences through learning of accumulated sensory-motor experiences. Yamashita \& Tani \cite{yamashita} and Noda et al. \cite{ogata} showed that a set of skilled behaviors like manipulating objects can be learned for robust generation using the predictive mechanism in this framework. Hwang et al. \cite{hwangsystem} showed that imitation learning using pixel level dynamic vision can be performed successfully by using predictive coding type deep visuomotor deep RNN model. Although the current application of the predictive coding is limited to simple prediction of action outcomes, it can be applied to more cognitively challenging problems involved with optimal action planning and their dynamic execution for achieving arbitrarily given goals. The current paper presents the first step toward such research goals by reporting a set of results from our robotic experiments.

The basic ideas and trails shown in the current study is briefly described in the following. The predictive coding scheme is implemented into a neural network model, referred to as predictive coding type deep visuomotor recurrent neural network model (P-DVMRNN). A real arm robot with vision is tutored for object-directed behavior generation tasks such as grasping an object for placing it on a goal target sheet. The tutoring is repeated for teaching a set of different trajectories dealt with large variation in positions such as for the object and the target goal sheet. The tutoring of each goal-directed trajectory for a particular task provides a robot with related multimodal perceptual experience consisting of pixel level vision and proprioception in terms of the joint angles which are extended in time as synchronized. 

A set of visuo-proprioceptive sequences obtained through tutoring of a particular task is used for off-line training of the P-DVMRNN model. P-DVMRNN model learns to regenerate each training trajectory by inferring the corresponding intention state. Here, the intention state which is represented by the internal neural activity in the model network encodes the way of the robot intending to interact with the environment.  It is noted that each intention state is self-determined in the course of learning. Consequently, after adequate training with good generalization it is expected that a causal mapping from the intention state space to the corresponding perceptual sequence space can be developed in the model network. After successful learning, the model network can generate mental image for visuo-proprioceptive sequence for the intention state inferred for the tutoring sequence. Moreover, it is assumed that an intention state located neighboring among those inferred in the training can generate analogous one by possibly interpolating those trained trajectories if generalization in learning can be done successfully. 

Let us consider further extension of the scheme to involve with goal-directed planning as the main objective in the current study. Suppose that goal state is given in terms of the corresponding perceptual state such as a visual frame image of a robot putting an object on a goal sheet. Then, the problem of planning is to infer the corresponding intention state which can achieve the specified perceptual state in the distal step by inversely applying the acquired causality between the intention to the perceptual sequence. Although it would be trivial to generate corresponding trajectories to a prior learned goal states, the same may not be assured for the case of unlearned goal states. We examine this issue by conducting robotic experiments by changing task difficulty. Although all the robot tasks considered in the current study might be relatively simple, our trial should be the first one for applying the predictive coding framework to the learning-based robot action planning by using deep learning scheme. The paper will focus on some difficulty we encountered in terms of generalization in planning and will discuss how the problem could be resolved by improving the scheme in future.

\section{Method}
\label{sec:arch}
In the predictive coding framework, all three processes of learning, recognition, and generation can be conducted by means of the prediction error minimization. Firstly, the learning is a process to map between intention states and the resultant perceptual sequences by self-determining the corresponding intention state for each sequence and connectivity weights for minimizing the prediction error. In the case of using RNN models for implementing the predictive coding scheme as like in the current study, the intention state can be represented by the initial states of internal neural units by utilizing the initial sensitivity characteristics of the RNN dynamics. Recognition is a process to infer inversely the corresponding intention state for a given target perceptual sequence. Finally, plan generation is to infer the corresponding intention state to achieve a goal state given at the distal step. The intentions state inferred is used to generate perceptual sequence reaching to the goal state. Next, we show how this predictive coding idea can be implemented in the current proposed neural network model.

\subsection{Neural Network Architecture}
\begin{figure}[thpb]
  \centering
  \includegraphics[scale=0.3]{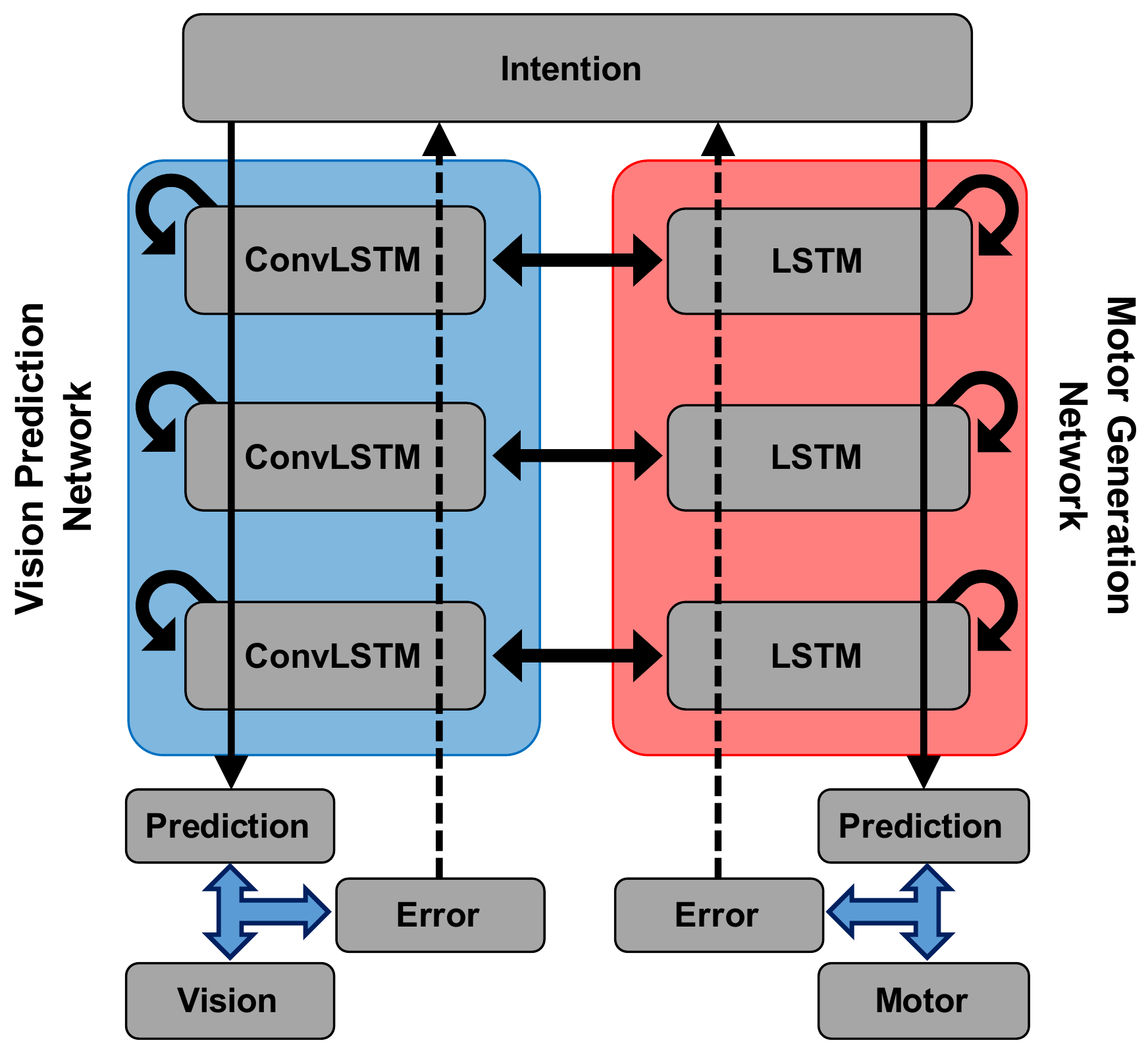}
  \caption{Overall network architecture. The proposed predictive coding type deep recurrent neural network model dealing with visuomotor sequences. }
  \label{arch}
\end{figure}

Our network architecture (shown in Fig. \ref{arch}) uses a recurrent neural network (RNN) based on predictive coding \cite{rao} capable of learning, generating, and recognizing multi-modality perceptual sequence inputs. The network has two closely related paths dedicated to processing visual input and motor joint angles respectively. At each time step, visual input in the form of a frame captured from an RGB camera is provided to the network, as well as the corresponding motor joints angles from a robot arm. The visual image and joint angles are fed as inputs to the lowest layer of the network and processed through three layers, then finally merged in the highest layer. The outputs predicting both the next visual input and joint angles are generated based on the internal neural activity in the lowest layer. Each layer is only connected to its neighbors (above/below) and the adjacent visual/motor counterpart. These structural characteristics enable network to process incoming data in an hierarchical manner \cite{hwangsystem,choi}. 

For the visual path, convolutional Long Short-term Memory (LSTM) networks \cite{convlstm} are used as the basic building blocks of the network in order to process spatial and temporal information simultaneously \cite{hwangsystem,jung,choi,convlstm}. In each path, there are two streams of information: top-down and bottom-up. The top-down connection projects the current prediction to the lower layers, while the bottom-up connection carries information from outside the network or errors between predictions and actual inputs. Top-down connections in vision path utilize convolutional operations and pooling, while the bottom-up connections are implemented as a transposed convolution \cite{deconv}. The size of the feature maps for each layer is designed to be half of the previous lower level. 

For the motor path, which operates on lower dimensional data compared to the visual path, LSTM is used. Similar to the feature maps in the visual path, the number of neurons in the motor path decreases along the hierarchy in the network. In order to improve learning with low dimensional data, sparse encoding is utilized \cite{hwangsystem,yamashita}. In this work, each motor joint is represented by a 10 dimensional sparsely encoded vector. 

The number of layers in both visual and motor pathways are identical, and as mentioned previously the layers at the same level in both visual and motor pathways are connected to each other horizontally. In Fig. \ref{arch}, the lateral connections at each level enable the whole network to exchange information between the two paths. Thus, this lateral connection is key for the network to closely couple the two modalities and maintains the link between a given visual input and motor joint angle. The network forms the common internal states of the highest layer by recognizing both visual input and current motor state from bottom-up connections. Since the raw visuomotor information is processed hierarchically through multiple layers from the lowest layer to highest layer, this internal state presents abstract information of the current environment as well as robot's state. Based on this abstract representation, the model is able to predict future visuomotor input by projecting it towards the lowest layer, generating a pixel-level visual prediction and a sparsely encoded joint angle prediction. 
A breakdown of each layer in the two network paths is shown in Table \ref{networkdetail}.

\begin{table}[]
\setlength\tabcolsep{5pt}
\centering
\caption{Breakdown of the size \& number of convolution feature maps (vision) and LSTM layers (motor) per layer}
\label{networkdetail}
\begin{tabular}{|c|c|c|c|c|c|}
\hline
                                                      & In/Out & L1       & L2       & L3     & L4                                                                       \\ \hline
\begin{tabular}[c]{@{}c@{}}Vision\\ Path\end{tabular} & 64$\times$64$\times$1  & 32$\times$32$\times$40 & 16$\times$16$\times$80 & 8$\times$8$\times$80 & \multirow{2}{*}{\begin{tabular}[c]{@{}c@{}}4$\times$4$\times$12\\ Shared\end{tabular}} \\ \cline{1-5}
\begin{tabular}[c]{@{}c@{}}Motor\\ Path\end{tabular}  & 40     & 1024       & 1024       & 16     &                                                                          \\ \hline
\end{tabular}
\end{table}

\subsection{Training}
During training, our model learns to predict/generate a set of training sequences by inferring the corresponding intention states represented by initial states (IS) of the internal units for each sequence as well as connectivity weights using back-propagation through time (BPTT) \cite{bptt} towards the direction of minimizing the prediction error.  
In this model, IS are self-determined for each training sequence. 

\subsection{Inference of intention for novel sequences}
With the trained network parameters (for e.g., weights and biases), it is expected that each training sequence can be regenerated using the corresponding IS in a closed-loop manner\footnote{Closed-loop: giving the previous time step's output as the current step's input, as opposed to Open-loop: input to each time step is given from ground truth data.}. 
When a novel perceptual sequence pattern is given to the learned network, it can be recognized by inferring the corresponding IS values by means of error regression (ER) scheme for minimizing the prediction error without changing connectivity weights. When the learning can be done with sufficient generalization, it is generally assumed that a novel sequence pattern which is similar to a particular trained one tends to be inferred with a similar IS value.

For example, consider a scenario in which we wish to find the IS (\(h_0\)) which encodes a given sequence in a simple RNN. Since we have no information of the actual IS, we set the current IS to a random value and start searching. In this search, we first generate a prediction output using the randomly set IS in a closed-loop manner as noted previously. Given the random IS, the output (\(O_{1:T}\)) of this process is unlikely to match our target sequence (\(T_{1:T}\)), producing a prediction error (\(E_{prediction}\)). 
\begin{equation}
	E_{prediction} = \sum_{t=1}^T ||O_t - T_t||^2
\end{equation}

This prediction error (\(E_{prediction}\)) is then back-propagated through time (BPTT) \cite{bptt}. Unlike training, during the ER process, model parameters such as weights and biases are left unchanged. Only the IS is optimized for prediction error minimization. This process is iterated multiple times until the predicted output follows the target sequence by minimizing prediction error. Once the optimal IS is found, the network is able to generate (or decode) the corresponding sequence. 



\subsection{Planning}
\begin{figure}[!t]
\centering
\subfloat[Given condition for planning]{\includegraphics[width=\linewidth]{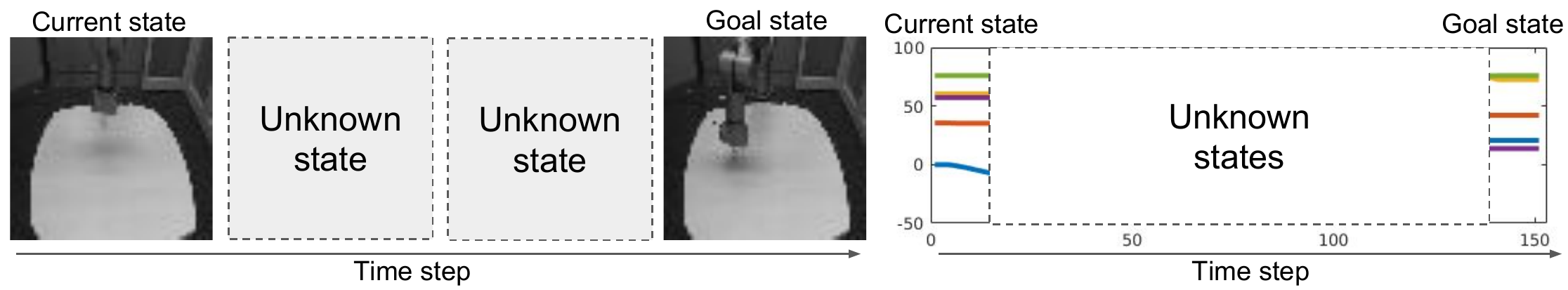}
\label{plan1}}
\hfil
\subfloat[Generated plan for vision and motor]{\includegraphics[width=\linewidth]{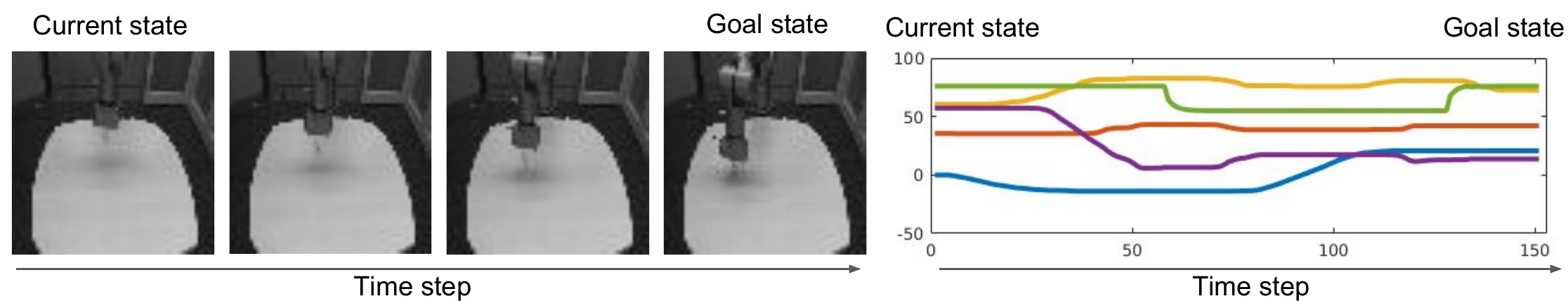}
\label{plan2}}
\caption{Planning by error regression. For both (a) and (b), left figures show visual data and right graphs present motor joint angles}
\label{initBias}
\end{figure}

This subsection describes how goal-directed action plans can be generated by extending the scheme of the ER. Let us suppose that the robot waits for a goal to be specified while staying at predefined home position posture. We consider that a goal state is given in terms of its corresponding perceptual state, i.e., visual state (\(V_{target, T}\)) and joint angle state (\(M^j_{target,T}\), where \(j\) is an index of joints and \(J\) is the number of joints, \(j \in J\)). Then the problem to solve is to generate an optimal visuomotor sequence which can rationally connect the perceptual state in the home posture in the initial step (\(V_{target,1}\) and \(M^j_{target,1}\)) and the one in the goal state. 
Fig. \ref{plan1} presents the available information for making plans and Fig. \ref{plan2} shows the generated visuomotor sequence connecting initial step and goal step.
Because the model network can generate various possible visuomotor sequences by changing the IS based on learning, it is considered that an optimal IS for generating such sequence can be searched by using the aforementioned ER scheme. A difference in the ER scheme for the plan generation is that the target perceptual states are given only partially, at the initial state and the end state. 
Thus, the prediction error which will be used to optimize an IS by ER is given as follows: 
\begin{equation} \label{eq:visionerror}
	\begin{split}
		E_{prediction} = &||V_{out,1} - V_{target,1}||^2 + ||V_{out,\hat{T}} - V_{target,T}||^2 
        \\&+ \sum_{j=1}^J KL(M^j_{out,1}|| M^j_{target,1})
        \\&+ \sum_{j=1}^J KL(M^j_{out,T}|| M^j_{target,T})
    \end{split}
\end{equation}
where \(V_{out,t}, V_{target,t}, M^j_{out,t}, M^j_{target,t}\) are the visual predicted output at step t, the visual target at step t, the \(j^{th}\) joint angle predicted output at step \(t\) and the \(j^{th}\) joint angle target at step \(t\) respectively. \(\hat{T}\) is a target step for the robot producing a target output. The time step at which the robot would achieve the given goal state is not known so it may be different from the ground truth target \(T\). Therefore, during the optimization process, \(\hat{T}\) is inferred. During the ER process, based on the current IS, a closed-loop prediction is generated until a predefined step \(T_{max}\) which is long enough to achieve the goal state. Then the ground truth visual target image \(V_{target,T}\) is compared against all generated prediction output frames (from \(V_{out,1}\) to \(V_{out,T_{max}}\)) and it generates errors for each step. In order to promote the model to achieve the goal faster, in a more optimized way, compensation value \(1.01^{t}\) is multiplied to the prediction error calculated for each time step. Among those predicted frames, the frame that has the smallest compensated error compared to \(V_{target,T}\) is set as \(V_{out,\hat{T}}\) and the respective output step \(\hat{T}\). 
As noted in Section \ref{sec:experiments}, this can result in a shorter sequence of steps to reach the goal.
In this work, as sparse coding is applied to the motor joint angles, Kullback-Leibler divergence (KL) is used to measure error between motor targets and outputs. 

When the IS for visual path is found by optimization, the robot is able to generate motor joint angles associated with each predicted image. As shown in the network architecture, the two modalities of vision and motor are correlated through lateral connections. Therefore, by searching optimal IS for one modality, it is possible to induce the other modality. Inducing motor joints angles is therefore possible by optimizing IS in the visual path and vice versa. 
The error from Equation ~\ref{eq:visionerror} describes the case when the both the visual and motor target is given. However, when only a target from one modality is given, eliminating the corresponding target term from Equation \ref{eq:visionerror} will yield a new prediction error. For example, in case the motor target is not given, the term \(\sum_{j=1}^J KLD(M^j_{out,T}|| M^j_{target,T})\) should be removed. 

\section{Experiments}
\label{sec:experiments}
In this section, we describe the experimental procedures and results obtained using an arm robot and camera connected to our network. Three tasks with different behavioral complexity were considered. The goals of the three tasks were, 1) reaching to a single point in the task space, 2) reaching to two points sequentially in the task space, and 3) grasping an object and putting it on a goal target sheet in the task space\footnote{Videos can be found at \url{https://sites.google.com/site/academicpapersubmission/p-dvmrnn}}. 

For the reaching tasks, our robot was configured to used 4 of its 7 joints, while in the grasping task, 5 joints including an end effector were used. The camera was in a fixed position facing the workspace and robot. While collecting training and testing data, both visual data and motor joint angles were sampled at 10Hz (for the grasping task, this was reduced to 2.5Hz). Each frame from the camera was resized to $64 \times 64$ pixels and 8 bit grayscale before being provided to the network.

Data collection was conducted in two phases: first, a human operator moved the robot by hand following a set of randomly generated positions. After the joint angles were recorded, the robot recreated the recorded trajectory and captured the video of the motion. For testing purposes, we only used the initial and final states of visuomotor trajectories from a test set and compared the prediction to the ground truth test trajectories. 
Because this data was generated by a human operator, it will naturally have noise and fluctuations. If our model is able to generalize the training trajectories to reach the goal state, it should be able to ignore the unnecessary pauses and fluctuations. Finally, both the pixel level vision and joint angles were normalized to $[-1, 1]$, and we utilized the Adam optimizer for training the network \cite{adam}. 

\subsection{Experiment 1: Reaching}
For the first experiment, a frame from the target vision data showing the last position of the robot arm is given as its goal state. To successfully accomplish this task, the network must generate a plausible prediction for both visual input and corresponding joint angles forming a trajectory of the arm robot. This experiment used a table (74cm x 74cm)
for the task space where 100 reaching trajectories for training were generated by randomly allocating the target position on the table. Testing was done on 40 randomly sampled positions that were not part of the training set.


\begin{figure}[!htbp]
  \centering
  \includegraphics[scale=0.8]{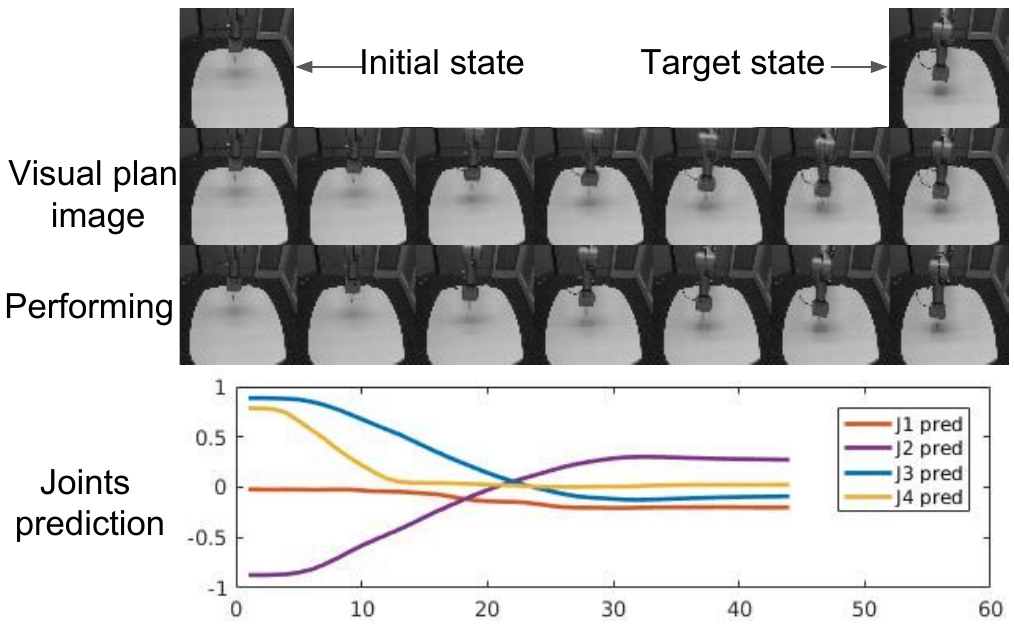}
  \caption{Example results from the reaching experiment. The upper row shows images of the initial and target visual states in the top level, visual image sequence generated for plan in the second level, and actual visual input perceived during execution of the plan in the bottom level. The lower row shows joint angles generated for the plan. Fig. 4 and Fig. 5 are presented in the similar manner.}
  \label{exp1-1}
\end{figure}

As described in equation \ref{eq:visionerror}, the IS of the network is optimized based on the errors between the visual predictions and visual targets given in the first and the last frames and then motor joint angles are generated based on the optimized IS. Fig. \ref{exp1-1} shows the results of this experiment. Although there is some visible blurring in prediction outputs (second row), the overall shape of the arm and its movement are maintained.
We also observe that the motor joint angles were successfully generated even though the target values of the joints were not provided to the network. 

As mentioned previously, since the training patterns were generated by a human operator, the trajectories are inconsistent. However, the robot reaches the goal state faster than the similar training trajectories. This suggests the model can generate more optimized trajectories that still reach the goal state, by generalizing training patterns.

To evaluate planning performance, we measured the distance between the final position reached based on the plan and the target position specified. Given the imprecision in the visual input, if the deviation at the end of the trajectory was less than 4\(cm\) (i.e., less than 3 pixels), the result was judged to be successful. Considering the overall size of robot arm (approximately 80\(cm\)), a 4\(cm\) error is believed to be reasonable. Over 40 test data points, the recorded average deviation was 2.6\(cm\) with a maximum deviation of 5.4\(cm\). The overall success rate of experiment 1 was 84\%.

Although some degree of position generalization was achieved as shown by the success rate of 84\% in the test generation, it was true that a relatively large amount of tutoring trajectories were used for the learning. Therefore, we examined how much the position generalization depended on the amount of tutoring trajectories. For this purpose, the same experiment was repeated by reducing number of tutoring trajectories, to 50 and 25. The result of the success rate in test generation is summarized in Table \ref{exp1_rate}. It can be seen that the success rate decreased significantly when the number of tutoring trajectories was reduced. It can be said that by using the current model, a reasonable success rate in test generation requires a relatively large amount of tutoring data of around 100 trajectories even for a relatively simple task as like the current one. 

\begin{table}[]
\centering
\caption{Success rate for experiment 1 with varying training set sizes}
\label{exp1_rate}
\begin{tabular}{|c|c|c|c|}
\hline
\begin{tabular}[c]{@{}c@{}}Training set size\end{tabular} & 25 & 50 & 100 \\ \hline
\begin{tabular}[c]{@{}c@{}}Average deviation\end{tabular} & 5.3\(cm\) & 3.2\(cm\) & 2.6\(cm\) \\ \hline
\begin{tabular}[c]{@{}c@{}}Success rate\end{tabular}  & 45\%     & 70\%       & 84\%  \\ \hline
\end{tabular}
\end{table}



\subsection{Experiment 2: Reaching two points}
For the second experiment, we extended the first task by adding an intermediate target that the robot must touch before reaching the goal. The intermediate target was marked by a filled circle with a diameter of 12\(cm\). The goal state was given as the last visual frame showing the intermediate target marker and the arm in the final position. 
To accommodate the two distributions of locations, the task space was expanded to 100\(cm\) by 100\(cm\).

\begin{figure}[!htbp]
  \centering
  \includegraphics[scale=0.8]{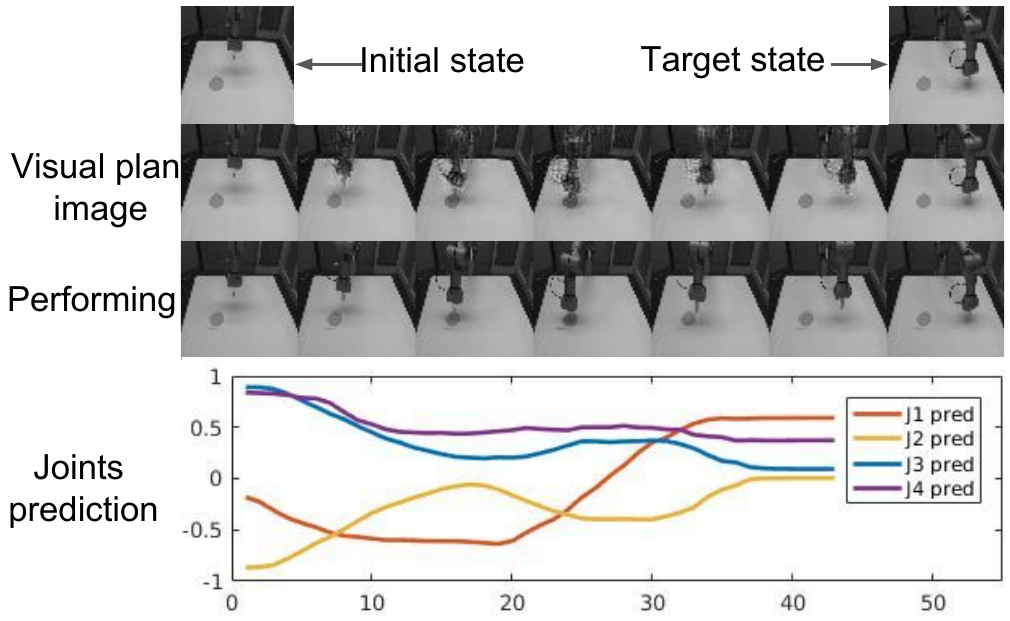}
  \caption{Results of touching two points experiment}
  \label{exp2}
\end{figure}

The task for the robot is to 1) touch the intermediate marker and then 2) move to the final position. For this task, if the robot touches a point within the marker and reaches the final position with a deviation of the end effector of less than 4\(cm\), the trial is regarded as successful. For training the network, 100 training sequences were collected. Fig. \ref{exp2} shows the target, predicted and actual visual frames as well as joint angles for one trial. The overall success rate was 75\% for this task.

\subsection{Experiment 3: Moving an object}
For the third experiment, we added an object to the task space for the robot to manipulate. The object and the target circle are placed in two randomly sampled locations as in experiment 2 and the task for the robot was to 1) grasp the object and 2) place it in the target circle. The object was a plastic cylinder with a diameter of 5\(cm\) and a height of 10\(cm\). The target circle and workspace were the same as in experiment 2. 100 training sequences and 50 test sequences are used. In testing, if the robot grasped the object and placed it upright anywhere within the target circle, it was considered a success.

\begin{figure}[!htbp]
  \centering
  \includegraphics[scale=0.8]{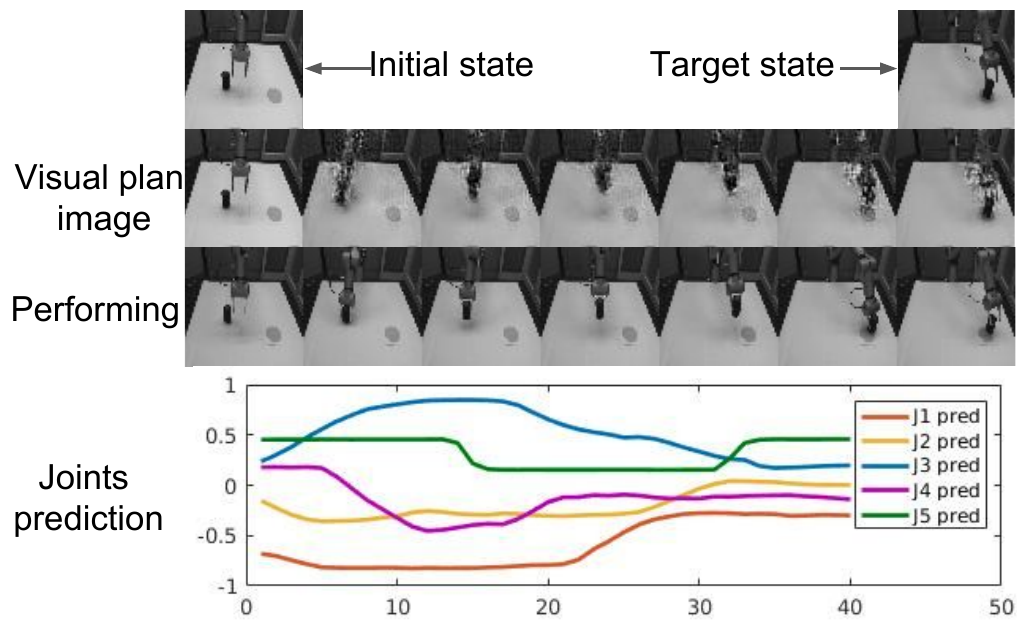} 
  \caption{Results of moving object experiment}
  \label{exp3}
\end{figure}

The difficulty of the task is considerably higher compared to the previous experiments, because the end effector must be moved accurately to grasp the object without sensory feedback. 
A successful trial is shown in Fig. \ref{exp3}. The overall success rate was 48\%. The challenge here was primarily the low resolution of visual input and resulting inaccurate predicted trajectories. Due to the size and shape of the the object and the end effector, any deviation greater than 2\(cm\) often resulted in failure to grasp the object. Despite this, we noted that once the robot grasped the object, it was able to successfully place the object in the target circle (94\% success rate, with an average deviation of 3\(cm\)).

In order to break down the performance in this task further, we considered the deviation from the center of the object to the center of where the end effector actuated. Allowing a 1 to 3 pixel error (1.3\(cm\) to 3.9\(cm\) deviation) in grasping, in line with previous tasks, the success rate was improved considerably as shown in Table \ref{exp3_rate}.

Additionally, we tested the ability of our model to produce one-step predictions. Unlike the previous tests, for one-step prediction the network observed ground truth visual data and motor joint angles after making a prediction at each timestep. This prediction scheme is employed in several other works \cite{rahmatizadeh,nagai,levine}. As the model receives sensory feedback, it yields better results than closed loop prediction. This difference is shown in Table \ref{exp3_rate} as closed loop prediction and open loop prediction respectively.

\begin{table}[]
\centering
\caption{Success rate for experiment 3 with varying amounts of error allowed in grasping}
\label{exp3_rate}
\begin{tabular}{|c|c|c|c|c|}
\hline
                                                                  & \multirow{2}{*}{\begin{tabular}[c]{@{}c@{}}Strict \\ grasping\end{tabular}} & \multicolumn{3}{c|}{Allowed error in grasping} \\ \cline{3-5} 
                                                                  &                                                                             & 1 pixel        & 2 pixels       & 3 pixels       \\ \hline
\begin{tabular}[c]{@{}c@{}}Closed loop \\Prediction\end{tabular} & 48\%                                                                        & 48\%           & 64\%          & 71\%          \\ \hline
\begin{tabular}[c]{@{}c@{}}Open loop \\Prediction\end{tabular}    & 74\%                                                                        & 74\%           & 88\%          & 93\%          \\ \hline
\end{tabular}
\end{table}


\section{Conclusion}
In this paper, we proposed a novel architecture for goal-directed action planning using a predictive coding type deep dynamic neural network. In the robot experiment, the network learned to generate a set of visuo-proprioceptive sequences by self-determining the corresponding intention state in terms of the IS for each sequence as well as connectivity weights in the whole network. After learning, the network was able to generate optimal visuomotor plans for the specified goal states by inferring the corresponding IS with some degree of generalization. Our experimental results have shown that our architecture can produce not only near future predictions (one-step ahead) as used in existing works, but also far future states (multi-step ahead) for both visual and motor modalities. However, due to restrictions in image resolution used in the vision network, the robot frequently failed in a grasping task that required precise positioning. This issue can be ameliorated somewhat by increasing image resolution or adding additional sensory input (for e.g., depth perception or tactile sensation) at the expense of increased computational cost.

A significant issue we observed was that to achieve fair generalization in learning and plan generation required a relatively large amount of training data. As shown with the first experimental task, the success rate in reaching the goal state was significantly reduced as number of training sequences was decreased. How can we solve this generalization problem? One possible solution may be to introduce a variational Bayes (VB) scheme to the model network \cite{kingma}. Recently, VB schemes have been introduced to several RNN models \cite{bengio,rezavb}. RNN models using a VB scheme show better generalization in learning by extracting probabilistic structures hidden in perturbed sequence data when the regularization term of controlling entropy in the neural activity is adequately tuned \cite{rezavb}. Examination of such models applied to learning-based goal-directed planning of robots is left for future study.


\addtolength{\textheight}{-12cm}   





\end{document}